\documentclass{esannV2}
\usepackage[dvips]{graphicx}
\usepackage[latin1]{inputenc}
\usepackage{amssymb,amsmath,array}
\usepackage{hyperref}
\usepackage{color}
\usepackage{subcaption}
\usepackage{tikz}
\usepackage{import}

\newcommand{\myheader}{Accepted as a conference paper at the \textit{European Symposium on Artificial Neural\\\hskip 1.3cm Networks, Computational Intelligence and Machine Learning (ESANN)} 2018}

\usepackage{atbegshi,picture}

\usepackage{datetime}            % used to put a date in the header

\usepackage{lastpage}            % organises page numbering

\newcommand{\myleftstd}{1.6in}

\ifthenelse{\isodd{\thepage}}
 {\newcommand{\myleftmargin}{\oddsidemargin+\myleftstd}}
 {\newcommand{\myleftmargin}{\evensidemargin+\myleftstd}}

\AtBeginShipoutNext { 
 \AtBeginShipoutUpperLeft{%
  \put(\dimexpr 0.2\myleftmargin\relax,-1.2cm){\parbox{\textwidth}{\footnotesize\sffamily\noindent{\centering \myheader \hfill}}}%
}}
\begin{document}
%style file for ESANN manuscripts
\title{Inferencing Based on Unsupervised Learning\\of Disentangled Representations}

%***********************************************************************
% AUTHORS INFORMATION AREA
%***********************************************************************
\author{Tobias Hinz and Stefan Wermter
%
% Optional short acknowledgment: remove next line if non-needed
\thanks{The authors gratefully acknowledge partial support from the German Research Foundation under project CML (TRR 169) and the European Union under project SECURE (No. 642667).}
%
% DO NOT MODIFY THE FOLLOWING '\vspace' ARGUMENT
\vspace{.3cm}\\
%
% Addresses and institutions (remove "1- " in case of a single institution)
Universit\"at Hamburg, Department of Informatics, Knowledge Technology \\
Vogt-Koelln-Str. 30, 22527 Hamburg, Germany \\
http://www.informatik.uni-hamburg.de/WTM/
}
%***********************************************************************
% END OF AUTHORS INFORMATION AREA
%***********************************************************************

\maketitle

%100 words
\begin{abstract}
Combining Generative Adversarial Networks (GANs) with encoders that learn to encode data points has shown promising results in learning data representations in an unsupervised way. We propose a framework that combines an encoder and a generator to learn disentangled representations which encode meaningful information about the data distribution without the need for any labels. While current approaches focus mostly on the generative aspects of GANs, our framework can be used to perform inference on both real and generated data points. Experiments on several data sets show that the encoder learns interpretable, disentangled representations which encode descriptive properties and can be used to sample images that exhibit specific characteristics.
\end{abstract}

\section{Introduction}
Learning meaningful representations of data is an important step for models to understand the world \cite{Bengio2013}. Recently, the Generative Adversarial Network (GAN) \cite{Goodfellow2014} has been proposed as a method that can learn characteristics of data distributions without the need for labels. GANs traditionally consist of a generator $G$, which generates data from randomly sampled vectors $Z$, and a discriminator $D$, which tries to distinguish generated data from real data $x$. During training, the generator learns to generate realistic data samples $G(Z)$, while the discriminator becomes better at distinguishing between the generated and the real data $x$. As a result, both the generator and the discriminator learn characteristics about the underlying data distribution without the need for any labels \cite{Radford2015}.
One desirable characteristic of learned representations is disentanglement \cite{Bengio2013}, which means that different parts of the representation encode different factors of the data-generating distribution. This makes representations more interpretable, easier to modify, and is a useful property for many tasks such as classification, clustering, or image captioning.

To achieve this, Chen et al. \cite{Chen2016} introduced a GAN variant in which the generator's input is split into two parts $z$ and $c$. Here, $z$ encodes unstructured noise while $c$ encodes meaningful, data-generating factors. Through enforcing high mutual information between $c$ and and the generated images $G(z, c)$ the generator is trained using the inputs $c$ as meaningful encodings for certain image characteristics. For example, a ten-dimensional categorical code for $c$ could represent the ten different digit classes in the MNIST data set. Since no labels are provided the generator has to learn by itself which image characteristics can be represented through $c$.
One drawback of this model is that the only way to perform inference, i.e.\ map real data samples into a (disentangled) representation, is to use the discriminator. However, there is no guarantee that the discriminator learns good representations of the data in general, as it is trained to discriminate between real and generated data and may therefore focus only on features that are helpful for discriminating these two, but are not necessarily descriptive of the data distribution in general \cite{Donahue2017}.
Zhang et al. \cite{Zhang2017} tried to enforce disentangled representations in order to improve the controllability of the generator. The latent representation is split up into two parts encoding meaningful information and unknown factors of variation. Two additional inference networks are introduced to enforce the disentanglement between the two parts of the latent representation. While this setup yields a better controllability over the generative process it depends on labeled samples for its training objective and can not discover unknown data-generating factors, but only encodes known factors of variation (obtained through labels) in its disentangled representation.

Donahue et al. \cite{Donahue2017} and Dumoulin et al. \cite{Dumoulin2017} introduced an extension which includes an encoder $E$ that learns the encodings of real data samples. The discriminator gets as input both the data sample $x$ (either real or generated) and the according representation (either $Z$ or $E(x)$) and has to classify them as either coming from the generator or the encoder. The generator and the encoder try to fool the discriminator into misclassifying the samples. As a result, the encoder $E$ learns to approximate the inverse of the generator $G$ and can be used to map real data samples into representations for other applications. However, in these approaches the representations follow a simple prior, e.g.\ a Gaussian or uniform distribution, and do not exhibit any disentangled properties.

Our model, the Bidirectional-InfoGAN, integrates some of these approaches by extending traditional GANs with an encoder that learns disentangled representations in an unsupervised setting. After training, the encoder can map data points to meaningful, disentangled representations which can potentially be used for different tasks such as classification, clustering, or image captioning. Compared to the InfoGAN \cite{Chen2016} we introduce an encoder to mitigate the problems of using a discriminator for both the adversarial loss and the inference task. Unlike the Structured GAN \cite{Zhang2017} our training procedure is completely unsupervised, can detect unknown data-generating factors, and only introduces one additional inference network (the encoder). In contrast to the Bidirectional GAN \cite{Donahue2017, Dumoulin2017} we replace the simple prior on the latent representation with a distribution that is amenable to disentangled representations and introduce an additional loss for the encoder and the generator to achieve disentangled representations.
On the MNIST, CelebA \cite{Liu2015}, and SVHN \cite{Netzer2011} data sets we show that the encoder does learn interpretable representations which encode meaningful properties of the data distribution. Using these we can sample images that exhibit certain characteristics, e.g. digit identity and specific stroke widths for the MNIST data set, or different hair colors and clothing accessories in the CelebA data set.

\begin{figure}
\centering
\tikzset{every picture/.style={line width=0.2mm}}
\begin{tikzpicture}[scale=0.39]
% left rectangle and content
\node[] at (2.5,9.5) {features};
\path[draw=black,fill=blue!10,opacity=0.200,line width=1pt,rounded corners=0.5cm] (0.5,-2) rectangle (4.7,9);
\draw[fill=white,rounded corners=0.2cm] (0.6,1.5) rectangle (4.6,2.8) node[pos=.5] {$E(x)$};
\draw[fill=white,rounded corners=0.2cm] (0.6,-0.2) rectangle (4.6,1.1) node[pos=.5] {$E(G(z,c))$};
\draw[fill=gray,opacity=0.3,rounded corners=0.2cm] (0.6,6.65) rectangle (4.6,7.95) node[pos=.5,opacity=1] {$(z,c)$};
\draw[->] (4.6,7.3) -- (7,7.3);
\draw[->] (9,7.3) -- (10.8,7.3);

\draw  (18.5,5.75) ellipse (2.7 and 1) node[] {$G(z,c),(z,c)$};

% right rectangle and content
\node[] at (13,9.5) {data};
\path[draw=black,fill=green!5,line join=miter,line cap=butt,miter
    limit=4.00,line width=1pt,rounded corners=0.5cm] (10.8,-2) rectangle (15,9);
\draw[fill=gray,opacity=0.3,rounded corners=0.2cm] (10.9,1.5) rectangle (14.9,2.8) node[pos=.5,opacity=1] {$x$};
\draw[->] (10.9,2.15) -- (10.5, 2.15) -- (10.5, 1.7) -- (9,1.7);
\draw[fill=white,rounded corners=0.2cm] (10.9,-0.2) rectangle (14.9,1.1) node[pos=.5] {$G(z,c)$};
\draw[->] (10.9,0.45) -- (10.5, 0.45) -- (10.5, 0.9) -- (9,0.9);
\draw[fill=white,rounded corners=0.2cm] (10.9,6.65) rectangle (14.9,7.95) node[pos=.5] {$G(z,c)$};
\draw[->] (2.5,6.6) -- (2.5,5.8) -- (15.8,5.8);
\draw[] (13,6.6) -- (13,5.8);

%line above
\draw[] (13,2.8) -- (13,3.4);
\draw[->] (2.5,2.8) -- (2.5,3.4) -- (15.8,3.4) node[pos=.4,above,inner sep=0.5pt] {\footnotesize if image} node[pos=.4,below,inner sep=0.5pt] {\footnotesize is real};

%line below
\draw[] (13,-1) -- (13,-0.2);
\draw[->] (2.5,-0.2) -- (2.5,-1) -- (15.8,-1) node[pos=.4,above,inner sep=0.5pt] {\footnotesize if image} node[pos=.4,below,inner sep=0.5pt] {\footnotesize is generated};

\draw  (18.5,3.35) ellipse (2.7 and 1) node[] {$x, E(x)$};
\draw  (19.8,-1) ellipse (4 and 1) node[] {$G(z,c), E(G(z,c))$};
\draw[fill=yellow!30]  (27.5,-1) ellipse (1.9 and 0.9) node[] {$L_\text{I}(G,E)$} node[above,inner sep=11pt] {\footnotesize mutual information};

\draw[->] (23.8,-1) -- (25.6,-1);

%generator node
\draw[fill=red!10] (7,8.3) -- (8,8.3) -- (9,7.3) -- (8,6.3) -- (7,6.3) -- (7,8.3);
\node[] at (7.75,7.3) {$G$};

%encoder node
\draw[fill=red!10] (8,2.3) -- (9,2.3) -- (9,0.3) -- (8,0.3) -- (7,1.3) -- (8,2.3);
\node[] at (8.25,1.3) {$E$};
\draw[->] (7,1.3) -- (6,1.3) -- (6,2.1) -- (4.7,2.1);
\draw[->] (6,1.3) -- (6,0.5) -- (4.7,0.5);

%discriminator node
\draw[fill=red!10] (23.5,5.6) -- (22.5,5.6) -- (22.5,3.6) -- (23.5,3.6) -- (24.5,4.6) -- (23.5,5.6);
\node[] at (23.25,4.6) {$D$};
\draw[->] (21.2,5.75) -- (21.5,5.75) -- (21.5,5.2) -- (22.5,5.2);
\draw[->] (21.2,3.45) -- (21.5,3.45) -- (21.5,4.2) -- (22.5,4.2);

\draw[fill=yellow!30]  (28,4.6) ellipse (2.5 and 0.9) node[] {$V(D,G,E)$} node[above,inner sep=11pt] {\footnotesize adversarial cost};
\draw[->] (24.5,4.6) -- (25.5,4.6);

\end{tikzpicture}
\vspace{-0.5cm}
\caption{High-level overview of the Bidirectional-InfoGAN. The generator $G$ generates images from the vector $(z, c)$ and tries to fool the discriminator into classifying them as real. The encoder $E$ encodes images into a representation and tries to fool the discriminator $D$ into misclassifying them as fake if its input is a real image while trying to approximate $P(c\vert x)$ if its input is a generated image.}
\vspace{-0.3cm}
\label{fig:model_architecture}
\end{figure}
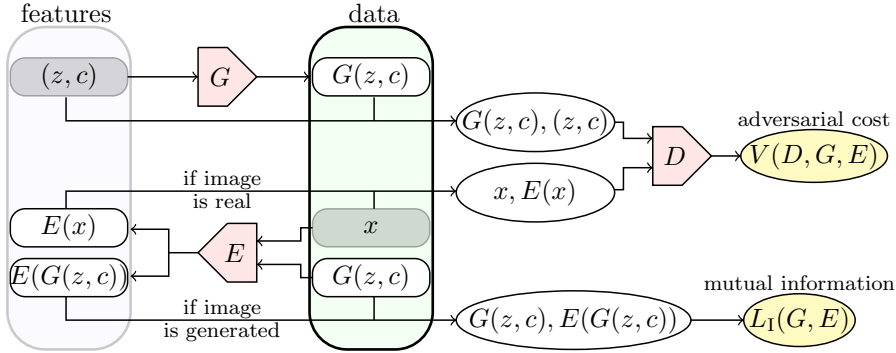

\section{Methodology}
Our model, shown in Fig.~\ref{fig:model_architecture}, consists of a generator $G$, a discriminator $D$, and an encoder $E$, which are implemented as neural networks. The input vector $Z$ that is given to the generator $G$ is made up of two parts $Z = (z, c)$. Here, $z$ is sampled from a uniform distribution, $z\sim U(-1, 1)$, and is used to represent unstructured noise in the images. On the other hand, $c$ is the part of the representation that encodes meaningful information in a disentangled manner and is made up of both categorical values $c_{\text{cat}}$ and continuous values $c_{\text{cont}}$. $G$ takes $Z$ as input and transforms it into an image $x$, i.e.\ $G: Z\rightarrow x$.

$E$ is a convolutional network that gets as input either real or fake images and encodes them into a latent representation $E: x\rightarrow Z$. $D$ gets as input an image $x$ and the corresponding representation $Z$ concatenated along the channel axis. It then tries to classify the pair as coming either from the generator $G$ or the encoder $E$, i.e.\ $D: Z\times x\rightarrow \{0,1\}$, while both $G$ and $E$ try to fool the discriminator into misclassifying its input. As a result the original GAN minimax game \cite{Goodfellow2014} is extended and becomes:
\[\underset{G, E}{\text{min}}\ \underset{D}{\text{max}}\ V(D, G, E) = \mathbb{E}_{x\sim P_{\text{data}}}[logD(x, E(x))] + \mathbb{E}_{Z\sim P_Z}[log(1-D(G(Z), Z))],\]
where $V(D, G, E)$ is the adversarial cost as depicted in Fig.~\ref{fig:model_architecture}.

\begin{figure}
\centering
\begin{subfigure}[b]{0.8\textwidth}
\begin{footnotesize}
\def\svgwidth{\linewidth}
\def\svgscale{0.5}
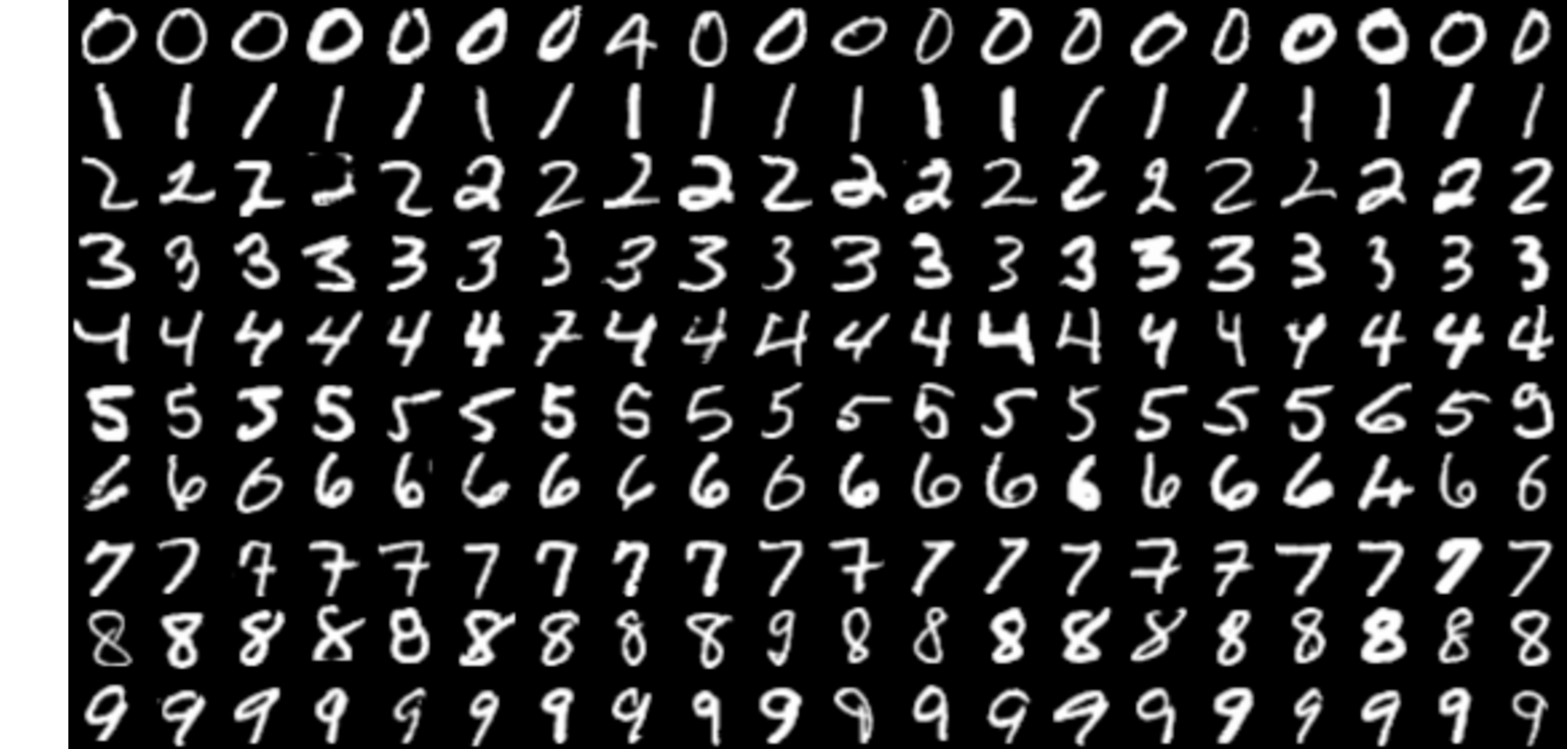
\end{footnotesize}
\vspace{-0.6cm}
\caption{}
\label{fig:mnist:discrete}
\end{subfigure}

\begin{subfigure}[b]{0.8\textwidth}
\begin{footnotesize}
\def\svgwidth{\linewidth}
\def\svgscale{0.5}
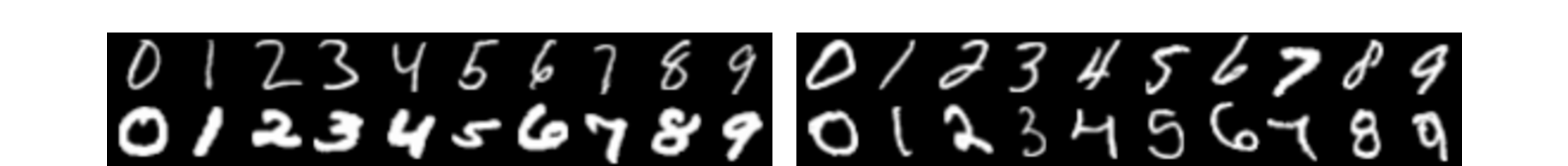
\end{footnotesize}
\vspace{-0.4cm}
\caption{}
\label{fig:mnist:continuous}
\end{subfigure}

\caption{Images sampled from the MNIST test set. (a) Each row represents one value of the ten-dimensional code $c_1$, which encodes different digits despite never seeing labels during the training process. (b) Images with maximum and minimum values for $c_2$ and $c_3$ for each categorical value from $c_1$.}
\vspace{-0.3cm}
\label{fig:mnist}
\end{figure}

In order to force the generator to use the information provided in $c$ we maximize the mutual information $I$ between $c$ and $G(z,c)$. Maximizing the mutual information directly is hard, as it requires the posterior $P(c\vert x)$ and we therefore follow the approach by Chen et al. \cite{Chen2016} and define an auxiliary distribution $E(c\vert x)$ to approximate $P(c\vert x)$. We then maximize the lower bound $L_I(G, E) = \mathbb{E}_{c\sim P(c), z\sim P(z), x\sim G(z,c)}[log\ E(c\vert x)]+H(c) \leq I(c;G(z,c))$, where $L_I(G, E)$ is the mutual information depicted in Fig.~\ref{fig:model_architecture}. For simplicity reasons we fix the distribution over $c$ and, therefore, the entropy term $H(c)$ is treated as a constant. In our case $E$ is the encoder network which gets images generated by $G$ as input and is trained to approximate the unknown posterior $P(c\vert x)$. For categorical $c_{\text{cat}}$ we use the softmax nonlinearity to represent $E(c_{\text{cat}}\vert x)$ while we treat the posterior $E(c_{\text{cont}}\vert x)$ of continuous $c_{\text{cont}}$ as a factored Gaussian. Given this structure, the minimax game for the Bidirectional-InfoGAN (BInfoGAN) is then
\[\underset{G, E}{\text{min}}\ \underset{D}{\text{max}}\ V_\text{BInfoGAN}(D, G, E) = V(D, G, E) - \lambda L_I(G, E)\]
where $\lambda$ determines the strength of the impact of the mutual information criterion $L_I$ and is set to $1.0$ in all our experiments.

\section{Experiments}
We perform experiments on the MNIST, the CelebA \cite{Liu2015}, and the SVHN \cite{Netzer2011} data set. While the final performance of the model is likely influenced by choosing the ``optimal'' characteristics for $c$ this is usually not possible, since we do not know all data-generating factors beforehand. When choosing the characteristics and dimensionality of the disentangled vector $c$ we therefore mostly stick with the values previously chosen by Chen et al. \cite{Chen2016}.
For further information on the network architectures and more examples of the learned characteristics on the different data sets see our Git: \url{https://github.com/tohinz/Bidirectional-InfoGAN}.

\begin{figure}
\centering
\begin{subfigure}[b]{0.8\textwidth}
\begin{footnotesize}
\def\svgwidth{\linewidth}
\def\svgscale{0.5}
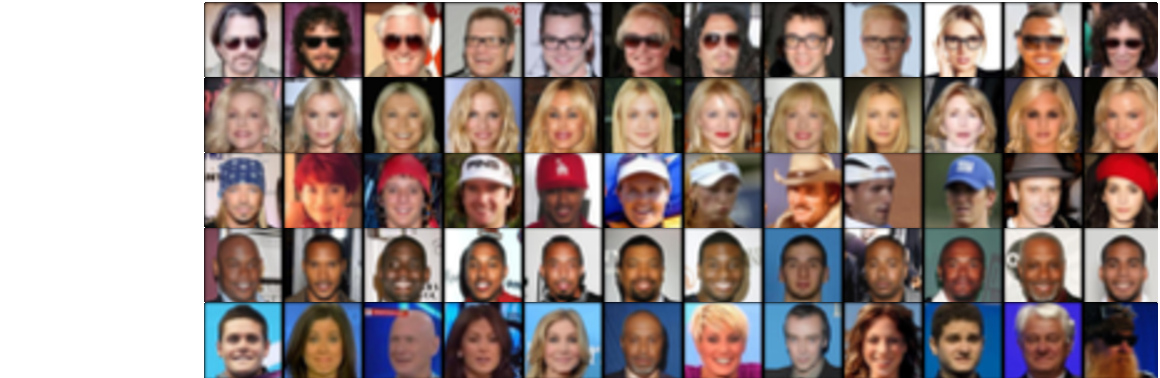
\end{footnotesize}
\vspace{-0.6cm}
\caption{}
\label{fig:celeba}
\end{subfigure}

\begin{subfigure}[b]{0.8\textwidth}
\begin{footnotesize}
\def\svgwidth{\linewidth}
\def\svgscale{0.5}
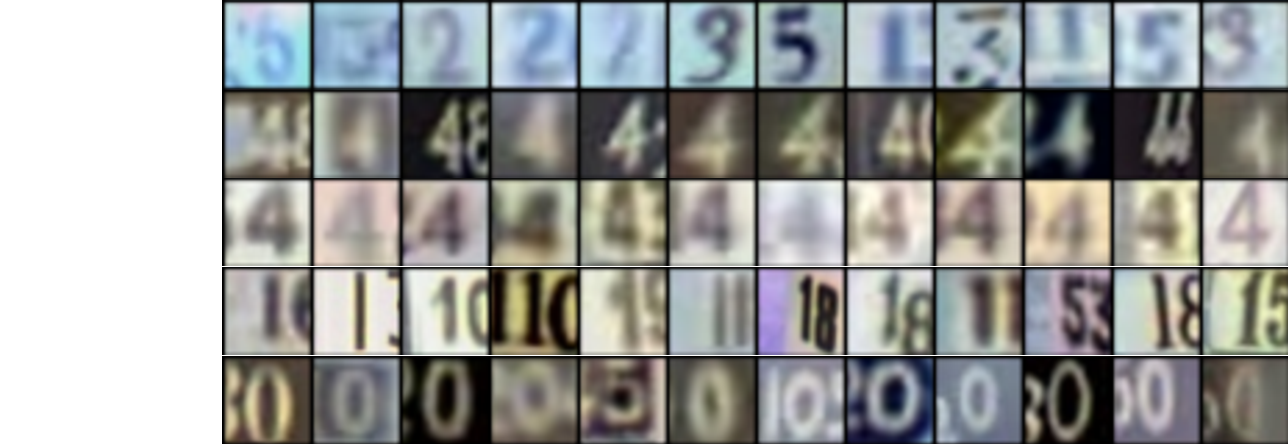
\end{footnotesize}
\vspace{-0.6cm}
\caption{}
\label{fig:svhn}
\end{subfigure}

\caption{Images sampled from the (a) CelebA and (b) SVHN test sets. Each row shows images sampled according to one specific categorical variable $c_{\text{cat}}$ which represents a learned characteristic.}
\vspace{-0.3cm}
\label{fig:celeba:svhn}
\end{figure}

On the MNIST data set we model the latent code $c$ with one categorical variable $c_1\sim Cat(K=10, p=0.1)$ and two continuous variables $c_2, c_3\sim U(-1,1)$. During the optimization process and without the use of any labels the encoder learns to use $c_1$ to encode different digit classes, while $c_2$ and $c_3$ encode stroke width and digit rotation. Fig.~\ref{fig:mnist:discrete} shows images randomly sampled from the test set according to the ten different categorical values. We can see that the encoder has learned to reliably assign a different categorical value for different digits. Indeed, by manually matching the different categories in $c_1$ to a digit type, we achieve a test set accuracy of 96.61\% ($\pm 0.32\%$, averaged over 10 independent runs)  without ever using labels during the training, compared to Chen et al. \cite{Chen2016} (unsupervised) with an accuracy of 95\%, and Zhang et al. \cite{Zhang2017} (semi-supervised, 20 labels) with an accuracy of 96\%. Fig.~\ref{fig:mnist:continuous} shows images sampled from the test set for different values of $c_2$ and $c_3$. We see that we can use the encodings from $E$ to now sample for digits with certain characteristics such as stroke width and rotation, even though this information was not explicitly provided during training.

On the CelebA data set the latent code is modeled with four categorical codes $c_1, c_2, c_3, c_4\sim Cat(K=10, p=0.1)$ and four continuous variables $c_5, c_6, c_7, c_8\sim U(-1,1)$. Again, the encoder learns to associate certain image characteristics with specific codes in $c$. This includes characteristics such as the presence of glasses, hair color, and background color and is visualized in Fig.~\ref{fig:celeba}.

On the SVHN data set we use the same network architecture and latent code representations as for the CelebA data set. Again, the encoder learns interpretable, disentangled representations encoding characteristics such as image background, contrast and digit type. See Fig.~\ref{fig:svhn} for examples sampled from the SVHN test set. These results indicate that the Bidirectional-InfoGAN is indeed capable of mapping data points into disentangled representations that encode meaningful characteristics in a completely unsupervised manner.

\section{Conclusion}
We showed that an encoder coupled with a generator in a Generative Adversarial Network can learn disentangled representations of the data without the need for any explicit labels. Using the encoder network we maximize the mutual information between certain parts of the generator's input and the images that are generated from it. Through this the generator learns to associate certain image characteristics with specific parts of its input. Additionally, the adversarial cost from the discriminator forces both the generator to generate realistic looking images and the encoder to approximate the inverse of the generator, leading to disentangled representations that can be used for inference.

The learned characteristics are often meaningful and humanly interpretable, and can potentially help with other tasks such as classification and clustering. Additionally, our method can be used as a pre-training step on unlabeled data sets, where it can lead to better representations for the final task. However, currently we have no influence over which characteristics are learned in the unsupervised setting which means that the model can also learn characteristics or features that are meaningless or not interpretable by humans. In the future, this can be mitigated by combining our approach with semi-supervised approaches, in which we can supply a limited amount of labels for the characteristics we are interested in to exert more control over which data-generating factors are learned while still being able to discover ``new'' generating factors which do not have to be known or specified beforehand.

% ****************************************************************************
% BIBLIOGRAPHY AREA
% ****************************************************************************
\begin{footnotesize}
\bibliographystyle{unsrt}
\bibliography{references.bib}
\end{footnotesize}

% ****************************************************************************
% END OF BIBLIOGRAPHY AREA
% ****************************************************************************

\end{document}